\documentclass[11pt, twocolumn]{article}

\usepackage{graphicx}
\graphicspath{ {./} }

\usepackage{amssymb}
\usepackage{amsmath}
\usepackage{adjustbox}
\usepackage{url}
\usepackage{hyperref} 
\usepackage{multirow}

\usepackage[left=1in,right=1in,top=1in,bottom=1in,%
footskip=.25in]{geometry}

\newcommand{\nostarnote}[1]{}
\newcommand{\baad}[1]{} 

\title{Enhancing Underwater Imagery using Generative Adversarial Networks}

\usepackage{authblk}
\author{Cameron Fabbri$^1$}
\author{Md Jahidul Islam$^2$}
\author{Junaed Sattar$^3$}
\affil{University of Minnesota, Minneapolis MN}
\affil{\{fabbr013$^1$, islam034$^2$, junaed$^3$\}@umn.edu}

\begin{document}

\twocolumn[
\begin{@twocolumnfalse}
	\maketitle
	\begin{abstract}
Autonomous underwater vehicles (AUVs) rely on a variety of sensors -- acoustic, inertial and visual -- for intelligent decision 
making. Due to its non-intrusive, passive nature, and high information content, vision is an attractive sensing modality, 
particularly at shallower depths. However, factors such as light refraction and absorption, suspended particles in the water, and 
color distortion affect the quality of visual data, resulting in noisy and distorted images. AUVs that rely on visual sensing thus 
face difficult challenges, and consequently exhibit poor performance on vision-driven tasks. This paper proposes a method to 
improve the quality of visual underwater scenes using Generative Adversarial Networks (GANs), with the goal of improving input to 
vision-driven behaviors further down the autonomy pipeline. Furthermore, we show how recently proposed methods are able to 
generate a dataset for the purpose of such underwater image restoration. For any visually-guided underwater robots, this 
improvement can result in increased safety and reliability through robust visual perception. To that effect, we present 
quantitative and qualitative data which demonstrates that images corrected through the proposed approach generate more visually 
appealing images, and also provide increased accuracy for a diver tracking algorithm.
	\end{abstract}
\end{@twocolumnfalse}
]

\section{Introduction}

Underwater robotics has been a steadily growing subfield of autonomous field robotics, assisted by the advent of novel platforms, 
sensors and propulsion mechanisms. While autonomous underwater vehicles are often equipped with a variety of sensors, visual 
sensing is an attractive option because of its non-intrusive, passive, and energy efficient nature. The monitoring of coral reefs 
\cite{shkurti2012multi}, deep ocean exploration \cite{whitcomb2000advances}, and mapping of the seabed~\cite{bingham2010robotic} 
are a number of tasks where visually-guided AUVs and ROVs (Remotely Operated Vehicles) have seen widespread use. Use of these 
robots ensures humans are not exposed to the hazards of underwater exploration, as they no longer need to venture to the depths 
(which was how such tasks were carried out in the past). Despite the advantages of using vision, underwater environments pose 
unique challenges to visual sensing, as light refraction, absorption and scattering from suspended particles can greatly affect 
optics. For example, because red wavelengths are quickly absorbed by water, images tend to have a green or blue hue to them. As 
one goes deeper, this effect worsens, as more and more red hue is absorbed. This distortion is extremely non-linear in nature, and 
is affected by a large number of factors, such as the amount of light present (overcast versus sunny, operational depth), amount 
of particles in the water, time of day, and the camera being used. This may cause difficulty in tasks such as segmentation, 
tracking, or classification due to their indirect or direct use of color. 

\begin{figure}[t]
\centering
\begin{tabular}{p{1.6cm} p{1.6cm} p{1.6cm} p{1.7cm}}
   \includegraphics[width=0.75in]{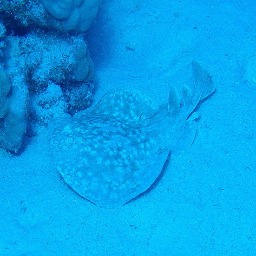} &
   \includegraphics[width=0.75in]{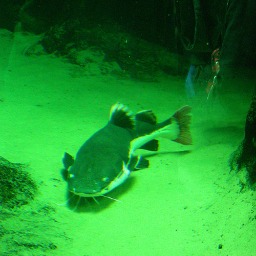} &
   \includegraphics[width=0.75in]{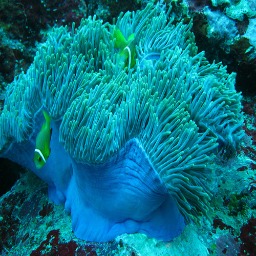} &
   \includegraphics[width=0.75in]{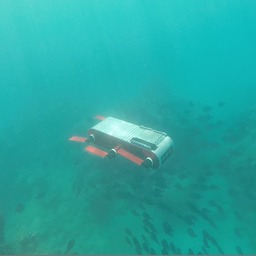} \\
   \includegraphics[width=0.75in]{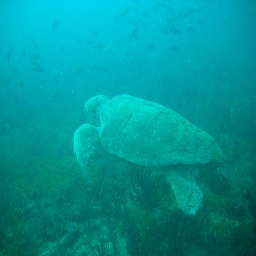} &
   \includegraphics[width=0.75in]{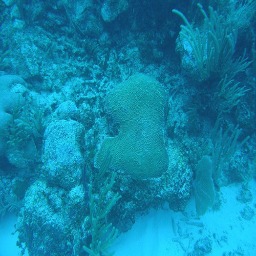} &
   \includegraphics[width=0.75in]{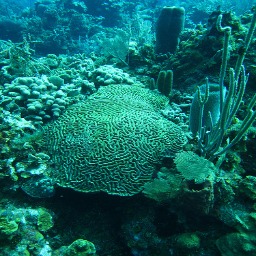} &
   \includegraphics[width=0.75in]{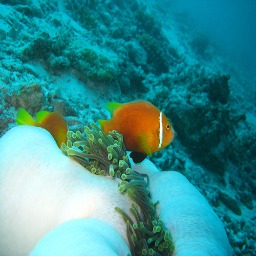} \\
\end{tabular}
\label{fig:samples}
\vspace{-2mm}
\caption{Sample underwater images with natural and man-made artifacts (which in this case is our underwater robot) displaying the 
diversity of distortions that can occur. With the varying camera-to-object distances in the images, the distortion and loss of 
color varies between the different images.}
\end{figure}

As color and illumination begin to change with depth, vision-based algorithms need to be generalizable in order to work within 
the depth ranges a robot may operate in. Because of the high cost and difficulty of acquiring a variety of underwater data to 
train a visual system on, as well as the high amount of noise introduced, algorithms may (and do) perform poorly in these 
different domains. Figure~\ref{fig:samples} shows the high variability in visual scenes that may occur in underwater environments. 
A step towards a solution to this issue is to be able to restore the images such that they appear to be above water, \emph{i.e.}, 
with colors corrected and suspended particles removed from the scene. By performing a many-to-one mapping of these domains from 
underwater to not underwater (what the image would look like above water), algorithms that have difficulty performing across 
multiple forms of noise may be able to focus only one clean domain.

Deep neural networks have been shown to be powerful non-linear function approximators, especially in the field of vision 
\cite{krizhevsky2012imagenet}. Often times, these networks require large amounts of data, either labeled or paired with
ground truth. For the problem of automatically colorizing grayscale images \cite{zhang2016colorful}, paired training
data is
readily available due to the fact that any color image can be converted to black and white. However, underwater images distorted 
by either
color or some other phenomenon lack ground truth, which is a major hindrance towards adopting a similar approach for correction. 
This paper proposes a technique based on Generative Adversarial Networks (GANs) to improve the quality of visual underwater scenes 
with the goal of improving the performance of vision-driven behaviors for autonomous underwater robots.
We use the recently proposed CycleGAN~\cite{zhu2017unpaired} approach, which learns to translate an image from any arbitrary 
domain $X$ to another arbitrary domain $Y$ \textit{without} image pairs, as a way to generate a paired dataset.
By letting $X$ be a set of undistorted underwater images\nostarnote{need to say where we find the undistorted ones}, and
$Y$ be a set of distorted underwater images, we can generate an image that appears to be underwater while retaining
ground truth.

\section{Related Work}
\label{sec:related}

While there have been a number of successful recent approaches towards automatic colorization 
\cite{zhang2016colorful,iizuka2016let}, most are focused on the task of converting grayscale images to color. Quite a few 
approaches use a physics-based technique to directly model light refraction~\cite{jordt2014underwater}. Specifically for
restoring color in underwater images, the work of~\cite{torres2005color} uses an energy minimization formulation using a Markov 
Random Field. Most similar to the work proposed in this paper is the recently proposed WaterGAN~\cite{li2017watergan}, 
which uses an adversarial approach towards generating realistic underwater images. Their generator model can be broken down into 
three stages: 1) Attenuation, which accounts for range-dependent attenuation of light. 2) Scattering, which models the haze 
effect caused by photons scattering back towards the image sensor and 3) Vignetting, which produces a shading effect on the 
image corners that can be caused by certain camera lenses. Differentiating from our work, they use a GAN for generating the 
underwater images and use strictly Euclidean loss for color correction, whereas we use a GAN for both. Furthermore, they require 
depth information during the training of WaterGAN, which can be often difficult to attain particularly for underwater 
autonomous robotic applications. Our work only requires images of objects in two separate domains (\emph{e.g.}, underwater 
and terrestrial) throughout the entire process.

Recent work in generative models, specifically GANs, have shown great success in areas such as inpainting 
\cite{pathak2016context}, style transfer \cite{Gatys_2016_CVPR}, and image-to-image translation 
\cite{isola2016image,zhu2017unpaired}. This is primarily due to their ability to provide a more meaningful loss than simply the 
Euclidean distance, which has been shown to produce blurry results. In our work, we structure the problem of estimating 
the true appearance of underwater imagery as a paired image-to-image translation problem, using Generative Adversarial Networks 
(GANs) as our generative model (see Section~\ref{sec:gans} for details). Much like the work of \cite{isola2016image}, we 
use image pairs from two domains as input and ground truth. \nostarnote{Somewhere I think we should mention that unpaired image 
to 
image translation is more difficult, and that CycleGAN is a good way to generate a dataset in such a way to do this with image 
pairs.}

\section{Methodology}
\label{sec:methodology}
Underwater images distorted by color or other circumstances lack ground truth, which is a necessity for previous colorization
approaches. Furthermore, the distortion present in an underwater image is highly nonlinear; simple methods such as
adding a hue to an image do not capture all of the dependencies. We propose to use CycleGAN as a distortion model in
order to generate paired images for training. Given a domain of underwater images with no distortion, and a domain of
underwater images with distortion, CycleGAN is able to perform style transfer. Given an undistorted image, CycleGAN
distorts it such that it appears to have come from the domain of distorted images. These pairs are then used in our
algorithm for image reconstruction.

\subsection{Dataset Generation}
Depth, lighting conditions, camera model, and physical location in the underwater environment are all factors that affect the 
amount of distortion an image will be subjected to. Under certain conditions, it is possible that an underwater image may have 
very little distortion, or none at all. We let 
$I^C$ be an underwater image with no distortion, and $I^D$ be the same image with distortion. Our goal is to learn the function 
$f: I^D \rightarrow I^C$. Becasue of the difficulty of collecting underwater data, more often than not only $I^D$ or $I^C$ exist, 
but never both.

To circumvent the problem of insufficient image pairs, we use CycleGAN to generate $I^D$ from $I^C$, which gives us a 
paired dataset of
\vspace{2pt}
images. Given two datasets $X$ and $Y$, where $I^C \in X$ and $I^D \in Y$, CycleGAN learns a mapping $F: X 
\rightarrow Y$. Figure~\ref{fig:cgan_samples} shows paired samples generated from CycleGAN. From this paired dataset we train a 
generator $G$ to learn the function $f: I^D \rightarrow I^C$. It should be noted that during the training process of CycleGAN, 
it simultaneously learns a mapping $G: Y \rightarrow X$, which is similar to $f$. In Section~\ref{sec:experiments}, we compare 
images generated by CycleGAN with images generated through our approach.

\vspace{-3pt}

\subsection{Adversarial Networks}
\label{sec:gans}
\nostarnote{check this paragraph as I added some more text}In machine learning literature, Generative Adversarial Networks 
(GANs)~\cite{goodfellow2014generative} represent a class of generative models based on game theory in which a generator network 
competes against an adversary. From a classification perspective, the generator network $G$ produces instances which actively 
attempt 
to `fool' the discriminator network $D$. The goal is for the discriminator network to be able to distinguish between `true' 
instances coming from the dataset and `false' instances produced by the generator network.
In our case, conditioned on an image $I^D$, the generator is trained to produce an image to try and fool the 
discriminator, which is trained to distinguish between distorted and non-distorted underwater images. In the original
GAN formulation, our goal is to solve the minimax problem:

\begin{figure}
\centering
\begin{tabular}{p{1.6cm} p{1.6cm} p{1.6cm} p{1.6cm}}
   \includegraphics[width=0.75in]{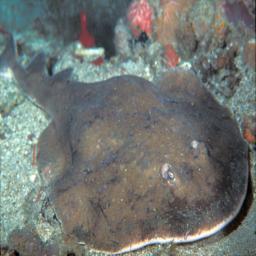} &
   \includegraphics[width=0.75in]{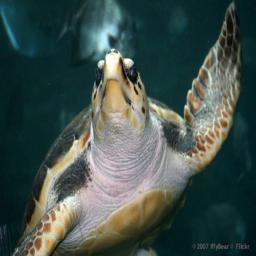} &
   \includegraphics[width=0.75in]{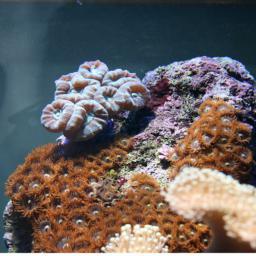} &
   \includegraphics[width=0.75in]{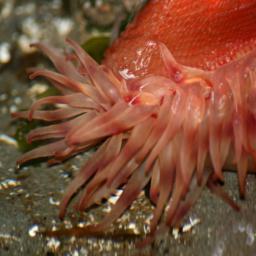} \\
   \includegraphics[width=0.75in]{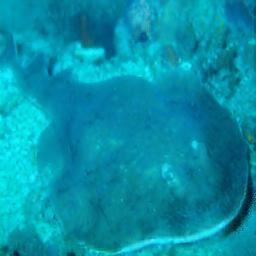} &
   \includegraphics[width=0.75in]{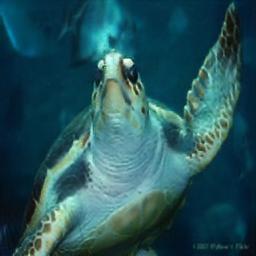} &
   \includegraphics[width=0.75in]{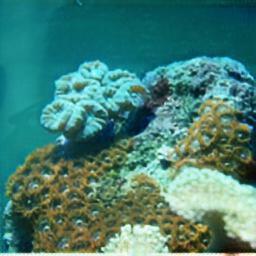} &
   \includegraphics[width=0.75in]{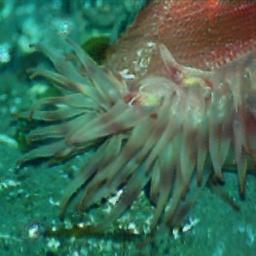} \\
\end{tabular}
\caption{Paired samples of ground truth and distorted images generated by CycleGAN. Top row: Ground truth.
Bottom row: Generated samples.}
\label{fig:cgan_samples}
\vspace{-4.4mm}
\end{figure}

\begin{equation}
\begin{aligned}
   \min\limits_{G}\max\limits_{D} \mathbb{E} & _{I^C \sim p_{train}(I^C)} [logD(I^C)] + \\
   \mathbb{E} & _{I^D \sim p_{gen}(I^D)}[log(1 - D(G(I^D)))]
\end{aligned}
\end{equation}

Note for simplicity in notation, we will further omit $I^C \sim p_{train}(I^C)$ and $I^D \sim p_{gen}(I^D)$. In this formulation, 
the discriminator is hypothesized as a classifier with a sigmoid cross-entropy loss function, which in practice may lead
to issues  such as the vanish gradient and mode collapse. As shown by \cite{arjovsky2017towards}, as
the discriminator improves, the gradient of the generator vanishes, making it difficult or impossible to train. Mode
collapse occurs when the generator ``collapses'' onto a single point, fooling the discriminator with only one instance.
To illustrate the effect of mode collapse, imagine a GAN is being used to generate digits from the MNIST~\cite{lecun2010mnist} 
dataset, but it only generated the same digit. In reality, the desired outcome would be to generate a diverse collection of all 
the digits. To this end, there have been a number of recent methods which hypothesize a different loss function for the 
discriminator
\cite{mao2016least,arjovsky2017wasserstein,gulrajani2017improved,zhao2016energy}. We focus on the Wasserstein GAN
(WGAN) \cite{arjovsky2017wasserstein} formulation, which proposes to use the Earth-Mover or \textit{Wasserstein-1}
distance $W$ by constructing a value function using the Kantorovich-Rubinstein duality \cite{villani2008optimal}.
In this formulation, $W$ is approximated given a set of $k$-Lipschitz functions $f$ modeled as neural networks. To
ensure $f$ is $k$-Lipschitz, the weights of the discriminator are clipped to some range $[-c, c]$. In our work, we
adopt the Wasserstein GAN with gradient penalty (WGAN-GP) \cite{gulrajani2017improved}, which instead of clipping
network weights like in \cite{arjovsky2017wasserstein}, ensures the Lipschitz constraint by enforcing a soft
constraint on the gradient norm of the discriminator's output with respect to its input. Following
\cite{gulrajani2017improved}, our new objective then becomes

\begin{equation}
\begin{aligned}
   \mathcal{L}_{WGAN}(G,D) = \mathbb{E} [D(I^C)] - \mathbb{E}  [D(G(I^D))] + \\
   \lambda_{GP} \mathbb{E}_{\hat{x} \sim \mathbb{P}_{\hat{x}}} [(|| \nabla_{\hat{x}} D(\hat{x})||_2 -1)^2 ]
\end{aligned}
\end{equation}


\noindent where $\mathbb{P}_{\hat{x}}$ is defined as samples along straight lines between pairs of points coming from
the true data distribution and the generator distribution, and $\lambda_{GP}$ is a weighing factor. In order to give $G$
some sense of ground truth, as well as capture low level frequencies in the image, we also consider the $L1$ loss

\begin{equation}
   \mathcal{L}_{L1} = \mathbb{E} [ || I^C - G(I^D) ||_1 ]
\end{equation}

\noindent Combining these, we get our final objective function for our network, which we call Underwater GAN (UGAN),

\begin{equation}
   \begin{aligned}
      \mathcal{L}_{UGAN}^* = \min\limits_{G}\max\limits_{D} \mathcal{L}_{WGAN}(G,D) + \lambda_{1} \mathcal{L}_{L1}(G)
   \end{aligned}
\end{equation}

\begin{figure*}[t]
\centering
\footnotesize
\begin{tabular}{p{0.1cm} p{1.6cm} p{1.6cm} p{1.6cm} p{1.6cm} p{1.6cm} p{1.6cm} p{1.6cm} p{1.6cm} }
   \rotatebox{90}{\: \, Original} &
   \includegraphics[width=0.7in]{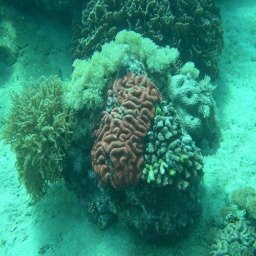} &
   \includegraphics[width=0.7in]{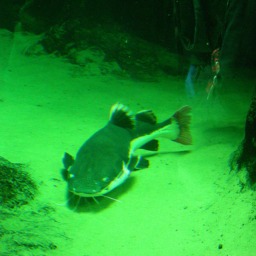} &
   \includegraphics[width=0.7in]{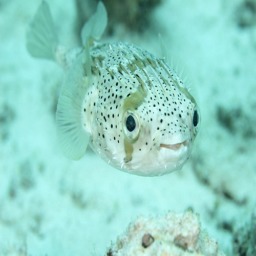} &
   \includegraphics[width=0.7in]{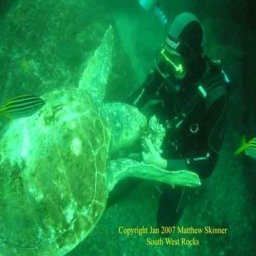} &
   \includegraphics[width=0.7in]{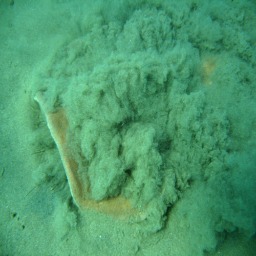} &
   \includegraphics[width=0.7in]{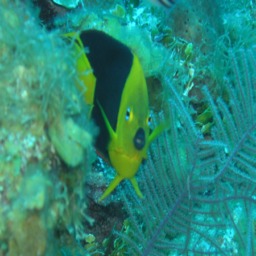} &
   \includegraphics[width=0.7in]{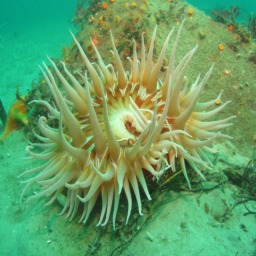} &
   \includegraphics[width=0.7in]{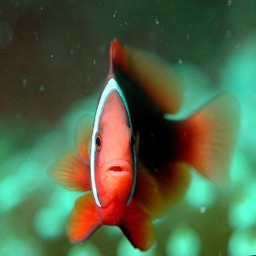} \\
   
   \rotatebox{90}{\textbf{\: \, UGAN}} &
   \includegraphics[width=0.7in]{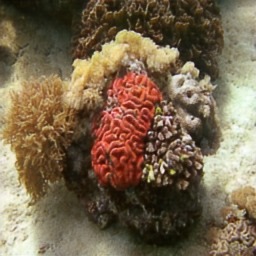} &
   \includegraphics[width=0.7in]{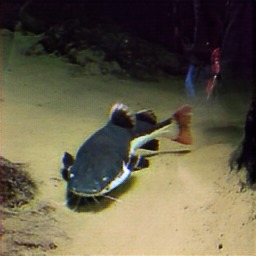} &
   \includegraphics[width=0.7in]{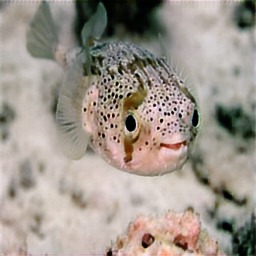} &
   \includegraphics[width=0.7in]{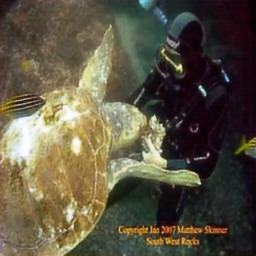} &
   \includegraphics[width=0.7in]{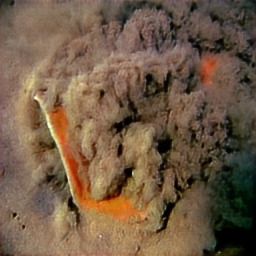} &
   \includegraphics[width=0.7in]{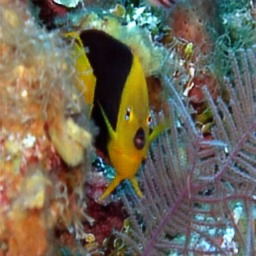} &
   \includegraphics[width=0.7in]{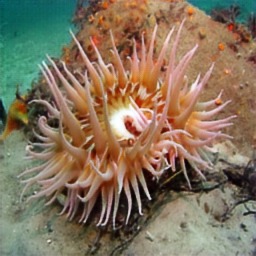} &
   \includegraphics[width=0.7in]{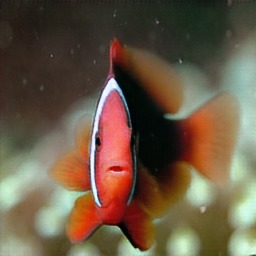} \\
   
   \rotatebox{90}{\textbf{\: UGAN-P}} &
   \includegraphics[width=0.7in]{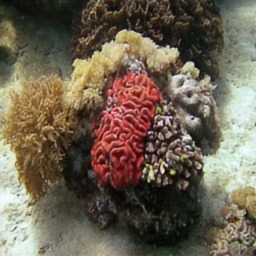} &
   \includegraphics[width=0.7in]{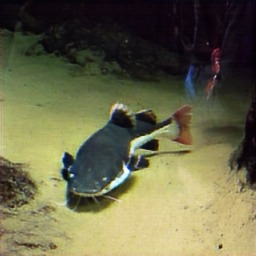} &
   \includegraphics[width=0.7in]{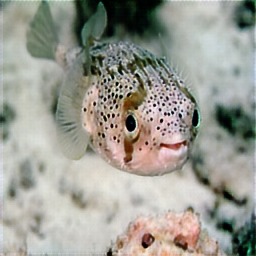} &
   \includegraphics[width=0.7in]{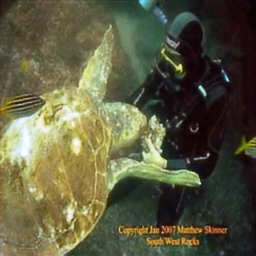} &
   \includegraphics[width=0.7in]{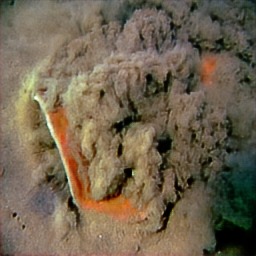} &
   \includegraphics[width=0.7in]{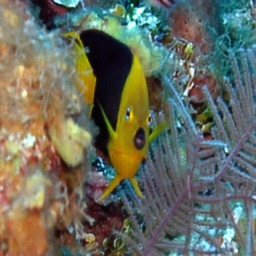} &
   \includegraphics[width=0.7in]{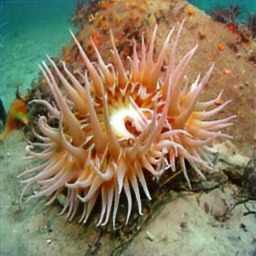} &
   \includegraphics[width=0.7in]{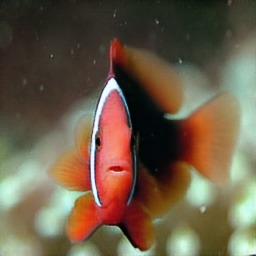} \\

\end{tabular}
\caption{Samples from our ImageNet testing set. The network can both recover color and also correct color if a small amount is 
present.}
\label{fig:test_samples}
\end{figure*}

\subsection{Image Gradient Difference Loss}
Often times generative models produce blurry images. We explore a strategy to sharpen these predictions by
directly penalizing the differences of image gradient predictions in the generator, as proposed by
\cite{mathieu2015deep}. Given a ground truth image $I^C$, predicted image $I^P = G(I^D)$, and $\alpha$ which is an integer 
greater than or equal to 1, the Gradient Difference Loss (GDL) is given by

\begin{equation}
   \begin{aligned}
      \mathcal{L}_{GDL}(I^C, I^P) = \\ \sum\limits_{i,j} || & I^C_{i,j} - I^C_{i-1,j}| - | I^P_{i,j} - I^P_{i-1,j}||^{\alpha} + \\
      || & I^C_{i,j-1} - I^C_{i,j}| - | I^P_{i,j-1} - I^P_{i,j}||^{\alpha}
   \end{aligned}
   \label{gdl_eq}
\end{equation}

\noindent In our experiments, we denote our network as UGAN-P when considering the GDL, which can be expressed as

\begin{equation}
   \begin{aligned}
      \mathcal{L}_{UGAN\scalebox{0.4}[1.0]{\( - \)}P}^* = \min\limits_{G}\max\limits_{D} \mathcal{L}_{WGAN}(G,D) + & \\
      \lambda_{1} \mathcal{L}_{L1}(G) + & \lambda_{2} \mathcal{L}_{GDL}
   \end{aligned}
\end{equation}

\subsection{Network Architecture}
Our generator network is a fully convolutional encoder-decoder, similar to the work of \cite{isola2016image}, which is
designed as a ``U-Net'' \cite{ronneberger2015u} due to the structural similarity between input and output.
Encoder-decoder networks downsample (encode) the input via convolutions to a lower dimensional embedding, in which
this embedding is then upsampled (decode) via transpose convolutions to reconstruct an image. The advantage of using
a ``U-Net'' comes from explicitly preserving spatial dependencies produced by the encoder, as opposed to relying on the
embedding to contain all of the information. This is done by the addition of ``skip connections'', which concatenate
the activations produced from a convolution layer $i$ in the encoder to the input of a transpose convolution layer
$n-i+1$ in the decoder, where $n$ is the total number of layers in the network. Each convolutional layer in our
generator uses kernel size $4 \times 4$ with stride 2. Convolutions in the encoder portion of the network are followed
by batch normalization \cite{pmlr-v37-ioffe15} and a leaky ReLU activation with slope $0.2$, while transpose
convolutions in the decoder are followed by a ReLU activation \cite{nair2010rectified} (no batch norm in the decoder).
Exempt from this is the last layer of the decoder, which uses a TanH nonlinearity to match the input distribution of
$[-1, 1]$. Recent work has proposed Instance Normalization \cite{ulyanov2016instance} to improve quality
in image-to-image translation tasks, however we observed no added benefit.

Our fully convolutional discriminator is modeled after that of \cite{radford2015unsupervised}, except no batch
normalization is used. This is due to the fact that WGAN-GP penalizes the norm of the discriminator's gradient with
respect to each input individually, which batch normalization would invalidate. The authors of
\cite{gulrajani2017improved} recommend layer normalization \cite{ba2016layer}, but we found no significant improvements.
Our discriminator is modeled as a PatchGAN \cite{isola2016image,li2016precomputed}, which discriminates at the level of
image patches. As opposed to a regular discriminator, which outputs a scalar value corresponding to real or fake, our
PatchGAN discriminator outputs a $32 \times 32 \times 1$ feature matrix, which provides a metric for high level
frequencies. 

\section{Experiments}
\label{sec:experiments}

\subsection{Datasets}
We used several subsets of Imagenet~\cite{deng2009imagenet} for training and evaluation of our methods. We also evaluate a 
frequency- and spatial-domain diver-tracking algorithm on a video of scuba divers taken from YouTube\texttrademark\ 
\footnote{https://www.youtube.com/watch?v=QmRFmhILd5o}. Subsets of 
Imagenet
containing underwater images were selected for the training of CycleGAN, and manually separated into two classes based on visual
inspection. We let $X$ be the set of underwater images with no distortion, and $Y$ be the set of underwater images with 
distortion. $X$
contained 6143 images, and $Y$ contained 1817 images. We then trained CycleGAN to learn the mapping $F: X \rightarrow Y$, such 
that images
from $X$ appeared to have come from $Y$. Finally, our image pairs for training data were generated by distorting all images in $X$ 
with
$F$. Figure~\ref{fig:cgan_samples} shows sample training pairs. When comparing with CycleGAN, we used a test set of 56 images 
acquired from
Flickr\texttrademark .

\begin{figure}[!ht]
\centering
\footnotesize
\begin{tabular}{p{1.6cm} p{1.6cm} p{1.6cm} p{1.6cm}}
   Original & CycleGAN & \textbf{UGAN} & \textbf{UGAN-P} \\
   \includegraphics[width=0.7in]{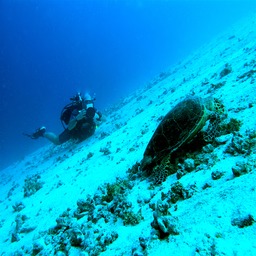} &
   \includegraphics[width=0.7in]{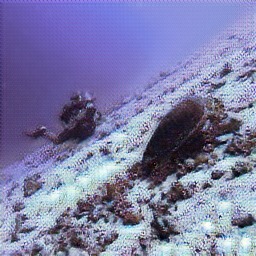} &
   \includegraphics[width=0.7in]{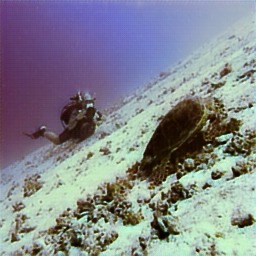} &
   \includegraphics[width=0.7in]{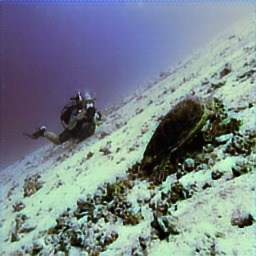} \\ [-1ex]
   \includegraphics[width=0.7in]{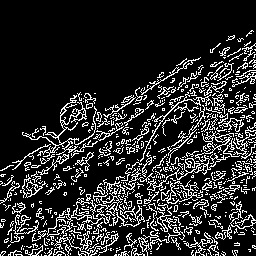} &
   \includegraphics[width=0.7in]{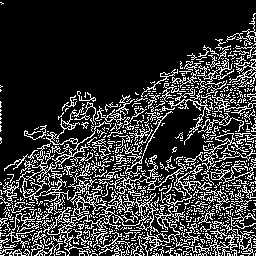} &
   \includegraphics[width=0.7in]{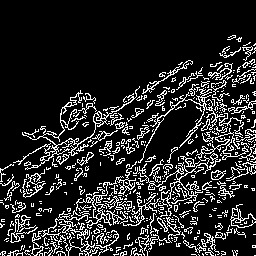} &
   \includegraphics[width=0.7in]{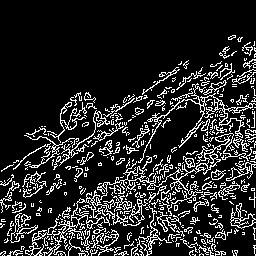} \\

   \includegraphics[width=0.7in]{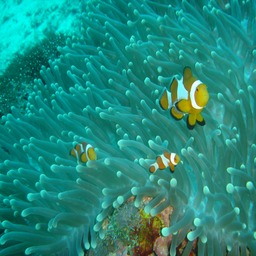} &
   \includegraphics[width=0.7in]{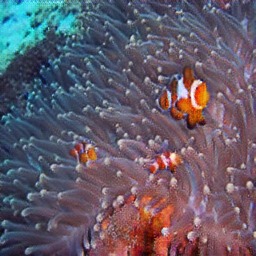} &
   \includegraphics[width=0.7in]{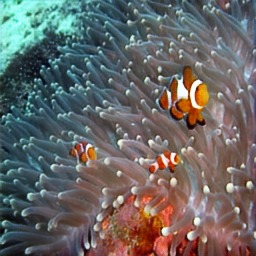} &
   \includegraphics[width=0.7in]{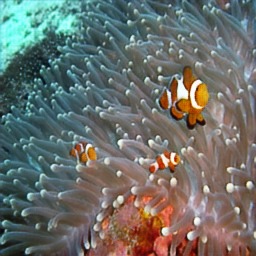} \\ [-1ex]
   \includegraphics[width=0.7in]{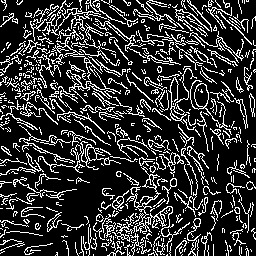} &
   \includegraphics[width=0.7in]{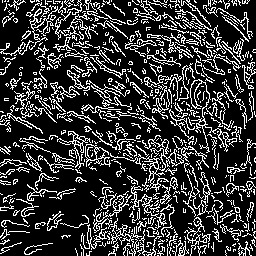} &
   \includegraphics[width=0.7in]{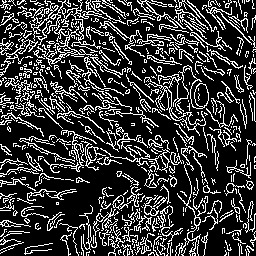} &
   \includegraphics[width=0.7in]{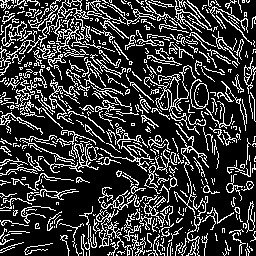} \\

   \includegraphics[width=0.7in]{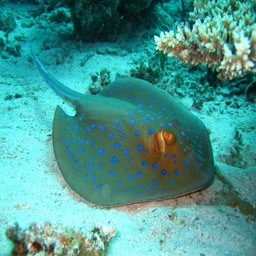} &
   \includegraphics[width=0.7in]{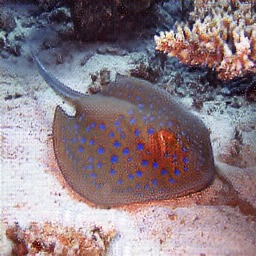} &
   \includegraphics[width=0.7in]{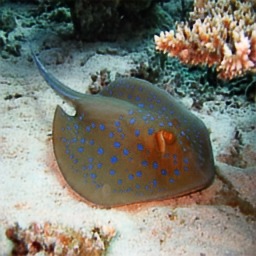} &
   \includegraphics[width=0.7in]{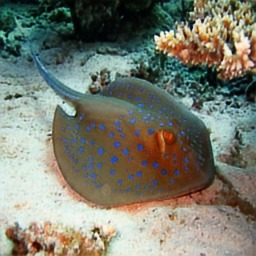} \\ [-1ex]
   \includegraphics[width=0.7in]{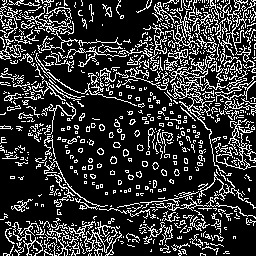} &
   \includegraphics[width=0.7in]{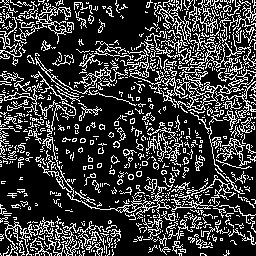} &
   \includegraphics[width=0.7in]{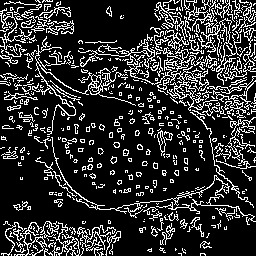} &
   \includegraphics[width=0.7in]{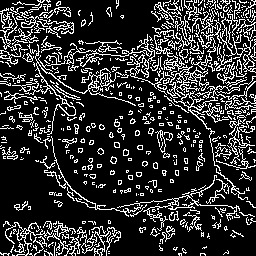} \\
   
   \includegraphics[width=0.7in]{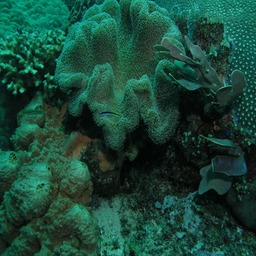} &
   \includegraphics[width=0.7in]{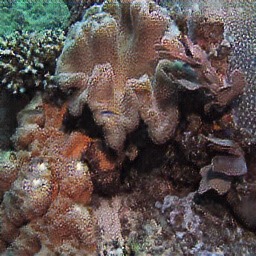} &
   \includegraphics[width=0.7in]{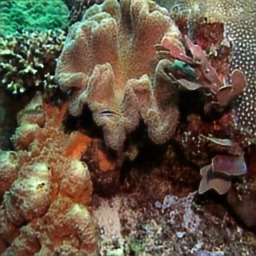} &
   \includegraphics[width=0.7in]{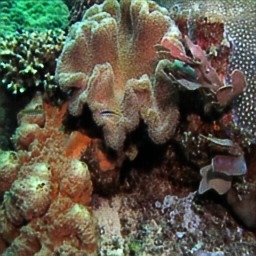} \\ [-1ex]
   \includegraphics[width=0.7in]{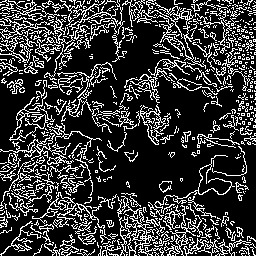} &
   \includegraphics[width=0.7in]{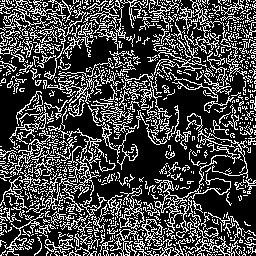} &
   \includegraphics[width=0.7in]{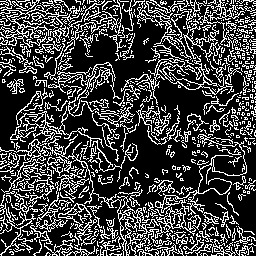} &
   \includegraphics[width=0.7in]{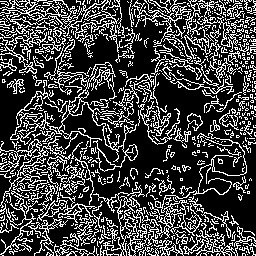} \\
\end{tabular}
\vspace{-2mm}
\caption{\small{Running the Canny Edge Detector on sample images. Both variants of UGAN contain less noise than CycleGAN,
and are closer in the image space to the original. For each pair, the top row is the input image, and bottom row
the result of the edge detector. The figure depicts four different sets of images, successively labeled A to D from top to 
bottom. See Table~\ref{tab:one}.}}
\label{fig:canny_samples}
\vspace{-5mm}
\end{figure}

\subsection{Evaluation}
We train UGAN and UGAN-P on the image pairs generated by CycleGAN, and evaluate on the images from the
test set, $Y$. Note that these images do not contain any ground truth, as they are original distorted images from
Imagenet. Images for training and testing are of size $256 \times 256 \times 3$ and normalized between $[-1, 1]$.
Figure ~\ref{fig:test_samples} shows samples from the test set. Notably, these images contain varying amounts of noise. Both UGAN 
and UGAN-P
are able to recover lost color information, as well as correct any color information this is present. 

While many of the distorted images contain a blue or green hue over the entire image space, that is not always the case.
In certain environments,
it is possible that objects close to the camera are undistorted with correct colors, while the background
of the image contains distortion. In these cases, we would like the network to only correct parts of the image that
appear distorted. The last row in Figure ~\ref{fig:test_samples} shows a sample of such an image. The orange of the clownfish is 
left
unchanged while the distorted sea anemone in the background has its color corrected.

For a quantitative evaluation we compare to CycleGAN, as it inherently learns an inverse mapping during the training of
$G: Y \rightarrow X$. We first use the Canny edge detector
\cite{canny1986computational}, as this provides a color agnostic evaluation of the images in comparison to ground truth.
Second, we compare local image patches to provide sharpness metrics on our images. Lastly, we show how an existing
tracking algorithm for an underwater robot improves performance with generated images.

\begin{figure*}[t]
\centering
\begin{tabular}{p{3.7cm} p{3.7cm} p{3.7cm} p{3.7cm}}
  
   \qquad \: \, ~ Original & \qquad \: \: CycleGAN & \qquad \: \: \, \textbf{UGAN} & \qquad \: \, \textbf{UGAN-P} \\

   \includegraphics[width=1.5in]{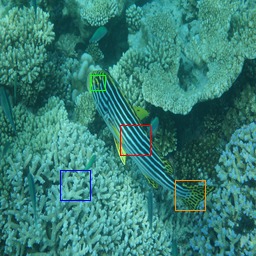} &
   \includegraphics[width=1.5in]{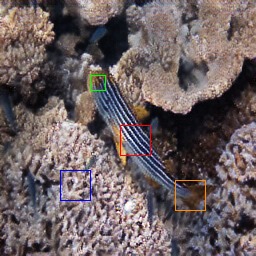} &
   \includegraphics[width=1.5in]{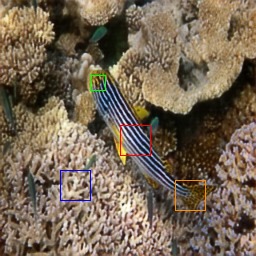} &
   \includegraphics[width=1.5in]{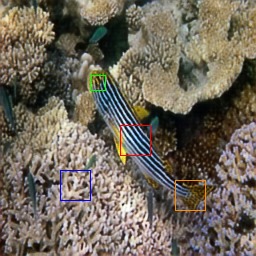} \\ [-2ex]
   
\end{tabular}
\end{figure*}

\begin{figure*}[t]
\begin{tabular}{p{1.6cm} p{1.7cm} p{1.6cm} p{1.7cm} p{1.6cm} p{1.7cm} p{1.6cm} p{1.9cm} }
   
   \includegraphics[width=0.7in]{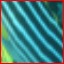} &
   \includegraphics[width=0.7in]{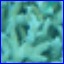} &
   \includegraphics[width=0.7in]{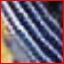} &
   \includegraphics[width=0.7in]{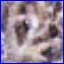} &
   \includegraphics[width=0.7in]{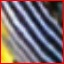} &
   \includegraphics[width=0.7in]{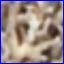} &
   \includegraphics[width=0.7in]{ugan_crop1} &
   \includegraphics[width=0.7in]{ugan_crop2} \\

   \includegraphics[width=0.7in]{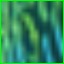} &
   \includegraphics[width=0.7in]{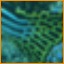} &
   \includegraphics[width=0.7in]{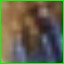} &
   \includegraphics[width=0.7in]{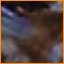} &
   \includegraphics[width=0.7in]{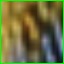} &
   \includegraphics[width=0.7in]{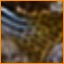} &
   \includegraphics[width=0.7in]{ugan_crop3} &
   \includegraphics[width=0.7in]{ugan_crop4} \\

\end{tabular}
\caption{Local image patches extracted for quantitative comparisons, shown in Tables~\ref{fig:gdl_tbl} and~\ref{fig:mean_tbl}. 
Each patch was resized to $64 \times 64$, but shown enlarged for viewing ability.}
\label{fig:zoom}
\end{figure*}

\subsection{Comparison to CycleGAN}
It is important to note that during the process of learning a mapping $F: X \rightarrow Y$, CycleGAN also learns a mapping $G: Y 
\rightarrow X$. Here we give a comparison to our methods. We use the Canny edge detector \cite{canny1986computational} to provide 
a color agnostic evaluation of the images, as the original contain distorted colors and cannot be compared back to as ground 
truth. Due to the fact that restoring color information should not alter the overall structure of the image, we measure the 
distance in the image space between the edges found in the original and generated images. Figure ~\ref{fig:canny_samples} shows 
the original images and results from edge detection. Table~\ref{tab:one} provides the measurements from Figure 
~\ref{fig:canny_samples}, as well as the average over our entire Flickr\texttrademark\ dataset. Both UGAN and UGAN-P are 
consistently closer in the image space to the original than that of CycleGAN, suggesting noise due to blur. Next, we evaluate this 
noise explicitly.

We explore the artifacts of content loss, as seen in Figure ~\ref{fig:zoom}. In particular, we compare local statistics of the 
highlighted image patches, where each image patch is resized to $64 \times 64$. We use the GDL \cite{mathieu2015deep} from 
(\ref{gdl_eq}) as a sharpness measure. A lower GDL measure implies a smoother transition between pixels, as a noisy image would 
have large jumps in the image's gradient, leading to a higher score.\nostarnote{I THINK. I would double check, bottom of page 4: 
https://arxiv.org/pdf/1511.05440.pdf.} As seen in Table \ref{fig:gdl_tbl}, the GDL is lower for both UGAN and UGAN-P. 
Interestingly, UGAN consistently has a lower score than UGAN-P, despite UGAN-P explicitly accounting for this metric in the 
objective function. Reasoning for this is left for our future work.

Another metric we use to compare image patches are the mean and standard deviation of a patch. The standard deviation gives us a 
sense of blurriness because it defines how far the data deviates from the mean.\nostarnote{ A lower standard deviation means that 
the data is closer to the mean . not sure if we exactly need to say that because I assume the reviewers will know what it is.} In 
the case of images, this would suggest a blurring effect due to the data being more clustered toward one pixel value. Table 
\ref{fig:mean_tbl} shows the mean and standard deviations of the RGB values for the local image patches seen in Figure 
\ref{fig:zoom}. Despite qualitative evaluation showing our methods are much sharper, quantitatively they show only slight 
improvement. Other metrics such as entropy are left as future work.

\begin{table}
\centering
\footnotesize
\caption{Distances in image space}
\begin{tabular}{| c | c | c | c |}
   \hline
   Row/Method & CycleGAN & \textbf{UGAN} & \textbf{UGAN-P} \\ \hline
   A          & 116.45 & 85.71  & 86.15  \\ \hline
   B          & 114.49 & 97.92  & 101.01 \\ \hline
   C          & 120.84 & 96.53  & 97.57  \\ \hline
   D          & 129.27 & 108.90 & 110.50 \\ \hline
   Mean       & 111.60 & 94.91  & 96.51 \\ \hline
\end{tabular}
\label{tab:one}
\end{table}

\begin{table}[ht]
\footnotesize
\centering
\caption{Gradient Difference Loss Metrics}
\begin{tabular}{| c | c | c | c | }
   \hline
   Method/Patch & CycleGAN & \textbf{UGAN} & \textbf{UGAN-P} \\ \hline
   Red    & 11.53 & 9.39 & 10.93  \\ \hline
   Blue   & 7.52  & 4.83 &  5.50\\ \hline
   Green  & 4.15  & 3.18 & 3.25 \\ \hline
   Orange & 6.72  & 5.65 & 5.79 \\ \hline
\end{tabular}
\label{fig:gdl_tbl}
\end{table}

\begin{table*}[ht]
\centering
\caption{Mean and Standard Deviation Metrics}
\begin{tabular}{| c | c | c | c | c | }
   \hline
   Method/Patch & Original & CycleGAN & \textbf{UGAN} & \textbf{UGAN-P} \\ 
\hline
   Red & 0.43 $\pm$ 0.23 & 0.42 $\pm$ 0.22 & 0.44 $\pm$ 0.23 & 0.45 $\pm$ 0.25 \\ \hline
   Blue & 0.51 $\pm$ 0.18 & 0.57 $\pm$ 0.17 & 0.57 $\pm$ 0.17 & 0.57 $\pm$ 0.17 \\ \hline
   Green & 0.36 $\pm$ 0.17 & 0.36 $\pm$ 0.14 & 0.37 $\pm$ 0.17 & 0.36 $\pm$ 0.17 \\ \hline
   Orange & 0.3 $\pm$ 0.15 & 0.25 $\pm$ 0.12 & 0.26 $\pm$ 0.13 & 0.27 $\pm$ 0.14 \\ \hline
\end{tabular}
\label{fig:mean_tbl}
\end{table*}

\subsection{Diver Tracking using Frequency-Domain Detection}
We investigate the frequency-domain characteristics of the restored images through a case-study of periodic motion tracking in 
sequence of images. Particularly, we compared the performance of Mixed Domain Periodic Motion (MDPM)- tracker 
\cite{islam2017mixed} on a sequence of images of a diver swimming in  arbitrary directions. MDPM tracker is designed for 
underwater robots to follow scuba divers by   tracking distinct frequency-domain signatures (high-amplitude spectra at $1$-$2$Hz) 
pertaining to human swimming. Amplitude spectra in frequency-domain correspond to the periodic intensity variations in image-space 
over time, which is often eroded in noisy underwater images \cite{shkurti2017underwater}.

\begin{figure*}[ht]
   \centering
   \begin{tabular}{p{3.7cm} p{3.7cm} p{3.7cm} p{3.7cm}}
      \includegraphics[width=1.5in]{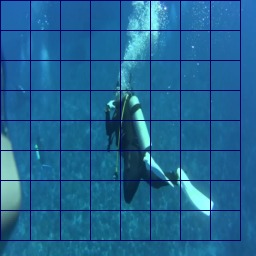} &
      \includegraphics[width=1.5in]{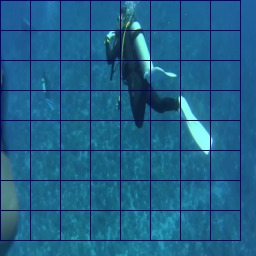} &
      \includegraphics[width=1.5in]{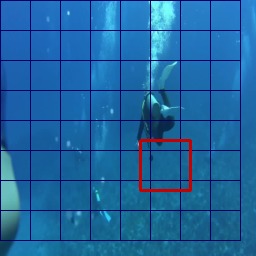} &
      \includegraphics[width=1.5in]{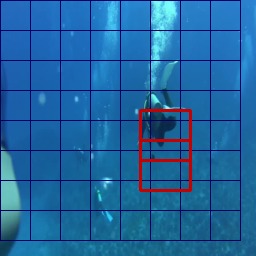} \\
      \includegraphics[width=1.5in]{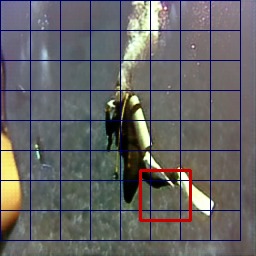} &
      \includegraphics[width=1.5in]{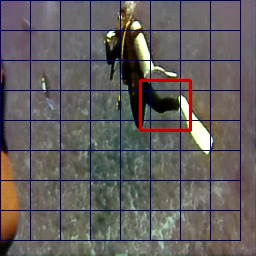} &
      \includegraphics[width=1.5in]{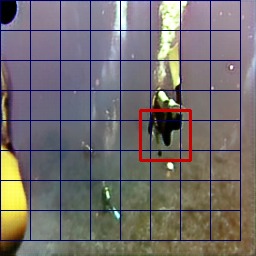} &
      \includegraphics[width=1.5in]{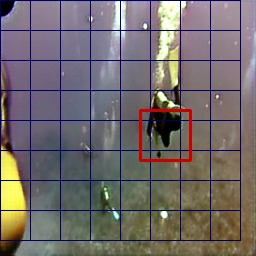} \\
   \end{tabular}
   \label{fig:mdpm}
   \vspace{4mm}

   \begin{tabular}{l|c|c|c|r|}
      \cline{2-5}
      &  Correct detection & Wrong detection & Missed detection & Total \# of frames\\ \hline  \cline{2-4}
      Real  &  42 & 14 & 444 & 500  \\ \hline
      Generated  &  147 & 24 & 329 & 500  \\ \hline
   \end{tabular}

   \caption{Performance of MDPM tracker \cite{islam2017mixed} on both real (top row) and generated (second row) images; the Table 
   compares the detection performance for both sets of images over a sequence of $500$ frames.   }
   \label{mdpmStuff}
\end{figure*}

Fig. \ref{mdpmStuff} illustrates the improved performance of MDPM tracker on generated images compared to the real ones. 
Underwater images often fail to capture the true contrast in intensity values between foreground and background due to low 
visibility. The generated images seem to restore these eroded intensity variations to some extent, causing much improved positive 
detection (a 350\% increase in correct detections) for the MDPM tracker.

\subsection{Training Details and Inference Performance}
In all of our experiments, we use $\lambda_{1} = 100$, $\lambda_{GP} = 10$, batch size of 32, and the Adam Optimizer 
\cite{kingma2014adam} with learning rate $1e-4$. Following WGAN-GP, the discriminator is updated $n$ times for every update of the 
generator, where $n = 5$. For UGAN-P, we set $\lambda_{2} = 1.0$ and $\alpha = 1$. Our implementation was done using the 
Tensorflow library \cite{abadi2016tensorflow}. \footnote{Code is available at 
\url{https://github.com/cameronfabbri/Underwater-Color-Correction}} All networks were trained from scratch on a GTX 1080 for 100 
epochs. Inference on the GPU takes on average $ 0.0138s$, which is about 72 Frames Per Second (FPS). On a CPU (Intel Core 
i7-5930K), inference takes on average $ 0.1244s$, which is about 8 FPS. In both cases, the input images have dimensions 
$256\times 256\times 3$. We find both of these measures acceptable for underwater tasks.


\section{Conclusion}

This paper presents an approach for enhancing underwater color images through the use of generative adversarial networks. We 
demonstrate the use of CycleGAN to generate dataset of paired images to provide a training set for the proposed restoration 
model. Quantitative and qualitative results demonstrate the effectiveness of this method, and using a diver tracking algorithm on 
corrected images of scuba divers show higher accuracy compared to the uncorrected image sequence.

Future work will focus on creating a larger and more diverse dataset from underwater objects, thus making the network more 
generalizable. Augmenting the data generated by CycleGAN with noise such as particle and lighting effects would improve 
the diversity of the dataset. We also intend to investigate a number of different quantitative performance metrics to evaluate 
our method.

\section*{Acknowledgment}
The authors are grateful to Oliver Hennigh for his implementation of the Gradient Difference Loss measure.
\newpage

\bibliography{cambibs}
\bibliographystyle{plain}

\end{document}